# Research on Limited Buffer Scheduling Problems in Flexible Flow Shops with Setup Times

**Zhonghua Han**

Department of Information and Control Engineering, Shenyang Jianzhu University, Shenyang, China
and
Department of Digital Factory, Shenyang Institute of Automation, Chinese Academy of Sciences, Shenyang, China
Email: xiaozhonghua1977@163.com

**Quan Zhang***

Department of Information and Control Engineering, Shenyang Jianzhu University, Shenyang, China
Email: zhangq0716@163.com
*Corresponding author

**Haibo Shi**

Department of Digital Factory, Shenyang Institute of Automation, Chinese Academy of Sciences, Shenyang, China
Email: hbshi@sia.cn

**Yuanwei Qi**

Department of Information and Control Engineering, Shenyang Jianzhu University, Shenyang, China
Email: qqyw2000@163.com

**Liangliang Sun**

Department of Information and Control Engineering, Shenyang Jianzhu University, Shenyang, China
Email: swinburnsun@163.com

**Abstract:** In order to solve the limited buffer scheduling problems in flexible flow shops with setup times, this paper proposes an improved whale optimization algorithm (IWOA) as a global optimization algorithm. Firstly, this paper presents a mathematic programming model for limited buffer in flexible flow shops with setup times, and applies the IWOA algorithm as the global optimization algorithm. Based on the whale optimization algorithm (WOA), the improved algorithm uses Levy flight, opposition-based learning strategy and simulated annealing to expand the search range, enhance the ability for jumping out of local extremum, and improve the continuous evolution of the algorithm. To verify the improvement of the proposed algorithm on the optimization ability of the standard WOA algorithm, the IWOA algorithm is tested by verification examples of small-scale and large-scale flexible flow shop scheduling problems, and the imperialist competitive algorithm (ICA), bat algorithm (BA), and whale optimization algorithm (WOA) are used for comparision. Based on the instance data of bus manufacturer, simulation tests are made on the four algorithms under variouis of practical evaluation scenarios. The simulation results show that the IWOA algorithm can better solve this type of limited buffer scheduling problem in flexible flow shops with setup times compared with the state of the art algorithms.

## 1  Introduction

Bus manufacturing workshop is a typical flexible flow-shop with the characteristics of multiple stages and multiple parallel machines. The bus has a long production cycle a lagre volume, so the production line can barely set the buffer with limited parking spaces. When the bus accesses the painting workshop, cleaning and device adjustments should be done on the painting machine if the properties such as the model and color are different from those of the previous bus processed by the painting machine. In this way, setup times will be added besides the standard processing time. Therefore, the scheduling problem of the bus manufacturer belongs to the limited buffer scheduling problems in flexible flow shops with setup times. Deng et al. presented an enhanced discrete artificial bee colony algorithm through a combined local search exploring both insertion and swap neighborhood to solve the flow shop scheduling with limited buffers (2016). Han et al. provided a solution to decide the reasonable size of the intermediate buffer, and proved a self-adaptive differential evolution algorithm as a practical solution (2018). Aiming at the limited buffer flow shop scheduling problem with minimum makespan as objective, a Hybrid Particle Swarm Optimization (HPSO) algorithm which combined Particle Swarm Optimization (PSO) with Iterated Greedy (IG) algorithm was proposed by Zhang and Chen (2012). A PSO-algorithm-based job scheduling method was presented by Han et al. to solve the production cost minimum problem (2012). Zhu et al. studied the multiple rules with game theoretic analysis for flexible flow shop scheduling problem with component altering times (2016). Multi-objective optimization for hybrid flow shop scheduling problem was solved by Han et al. with an improved NSGA-II algorithm (2017). Zeng et al. proposed an adaptive cellular automata variation particles warm optimization algorithm to solve the problems of flexible flow shop batch scheduling (2012). An improved imperialist competitive algorithm was proposed by Sun et al. as the searching mechanism with efficiencies for both convergence speed and global optimum searching (2017). Zhang and Pan presented a hybrid artificial bee colony algorithm combined with the Weighted Profile Fitting, which effectively solved the flow shop scheduling problem with limited buffers (2013). Zhang and Wong employed an enhanced version of ant colony optimization (E-ACO) algorithm to solve flow shop scheduling problems with setup times (2016). Zhong et al. proposed a hybrid evolutionary algorithm for solving the single-machine scheduling problem with setup times to minimize the total tardiness (2013). Rajesh studied the problem of scheduling a flow shop operating in a sequence-dependent setup time environment and adopted two heuristic algorithms to solve the problem (2014). To solve the scheduling problems of re-entrant hybrid flowshop, Han et al. proposes the wolf pack based algorithm as a global optimization method (2018).

The survey of the literatures in recent years shows that the researchers mainly focused on the optimization algorithm improvement for limited buffer scheduling problems, and the current scheduling research with setup times mostly investigated the impact of setup times under the single condition on the processing time. However, very few work has been done for studing the scheduling problems with limited buffer and setup times under the conditions of dynamic composition of multiple properties. The limited buffer scheduling problem in flexible flow shops with setup times is a kind of extremely complicated scheduling optimization problem. How to avoid premature convergence while ensuring optimization speed is a challenge issue. On the other hand, the WOA algorithm is a new swarm intelligence optimization algorithm proposed by Mirjalili S et al. in 2016, whose bionics derives from the hunting behavior of humpback whales. The algorithm has been widely used in recent years. For example, AI-Zoubi et al. used the WOA algorithm to detect spam in the network and identify its sender (2018). Rewadkar et al. proposed a multi-objective auto-regressive WOA algorithm for traffic-aware routing in urban VANET, which has achieved good results (2018). Kaveh, A and Ghazaan, MI put forward an enhanced WOA algorithm for sizing optimization of skeletal structures (2018). The WOA algorithm has the characteristics of simpler parameters and faster search speed, which is very suitable for dealing with complex optimization problems. Therefore, this paper uses whale optimization algorithm to solve the limited buffer scheduling problem in flexible flow shops with setup times. To overcome WOA's shortcomings, such as falling into local extremum easlily, this paper improves WOA by introducing Levy flight and opposition-based learning strategy. And the improved WOA can avoid the premature convergence of the algorithm while ensuring the convergence speed.

## 2 Mathematical model for limited buffer scheduling problems in flexible flow shops with setup times

### 2.1 Problem description

As shown in Figure 1, the limited buffer scheduling problems in flexible flow shops with setup time studied in this paper can be described as: n jobs are required to be processed in m stages. Among the stages, one stage has at least two parallel machines. There is at least one machine for each stage, and the same job has the same processing time on any parallel machine in the same stage. Buffers with given capacity are present between stages. When the job is processed in the previous stage, it enters the buffer and joins the waiting list at the next stage. If the buffer is full and there is not free machine in the next stage, the completed job of the previous stage will stay on the current machine. At this time, the machine is unavailable and can not process other jobs until the buffer has available spaces for other jobs. When the job is assigned to the machine, the setup time is added besides the standard processing time if the property of the job dose not match that of the previous job. Knowing the standard processing time of the job at each stage, the online sequence of the job in the production process and the job's setup times, start time, and completion time at each stage are obtained through the scheduling, so as to achieve better scheduling results (Han et al.,2016).

**Figure 1** Mathematical model for limited buffer scheduling problems in flexible flow shops with setup times

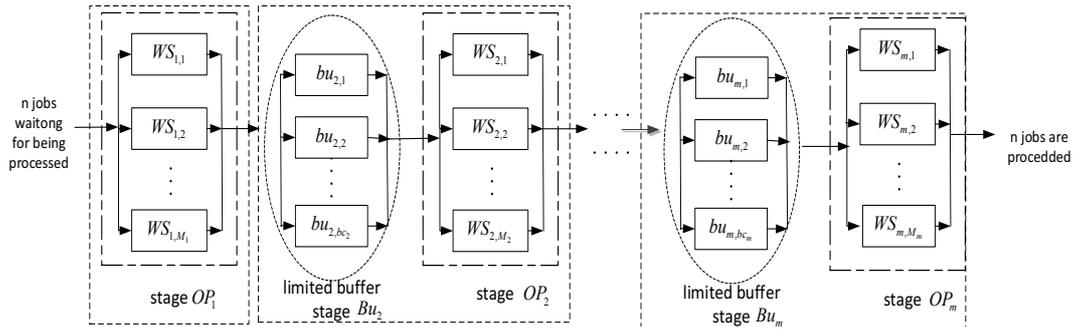

### 2.2 Parameters in the model

$n$ : Quantity of jobs to be scheduled;
$J_i$: Job $i$, $i$ $\{1,...,n\}$;
$m$: Quantity of stages;
$Oper_j$: Stage $j$, $j$ $\{1,...,m\}$;
$Bu_j$: The buffer of stage $Oper_j$, $j$ $\{2,...,m\}$;

$K_j$: Quantity of machines in the buffer $Bu_j$ at stage $Oper_j$, $j$ $\{2,...,m\}$;

$b_{j,k}$ : Machine $k$ in the buffer $Bu_j$, $j$ $\{2,...,m\}$, $k$ $\{1,...,K_{j,a}\}$;

$WA_j(t)$: At time $t$, the waiting sequences in the buffer $Bu_j$, $j$ $\{2,...,m\}$;

$card(WA_j(t))$: At time $t$, the quantity of jobs in the waiting processing sequence $WA_j$;

$M_j$: Quantity of machines in each stage, $j \in \{1,...,m\}$;

$WS_{j,l}$: Machine $l$ of stage $Oper_j$, $j \in \{1,...,m\}$, $l \in \{1,...,M_j\}$;

$S_{i,j}$: The start time to process job $J_i$ at stage $Oper_j$;

$C_{i,j}$: The completion time to process job $J_i$ at stage $Oper_j$;

$Tb_{i,j}$: The standard processing time of job $J_i$ at stage $Oper_j$, $j \in \{1,...,m\}$;

$Te_{i,j}$: The entry time of job $J_i$ into buffer $Bu_j$, $j \in \{2,...,m\}$;

$Tl_{i,j}$: The departure time of job $J_i$ out of buffer $Bu_j$, $j \in \{2,...,m\}$;

$To_{i,j}$: The departure time of job $J_i$ on its machine after job $J_i$ is processed at stage $Oper_j$, $j \in \{1,...,m\}$;

$Tw_{i,j}$: The waiting time of job $J_i$ in buffer $Bu_j$, $j \in \{2,...,m\}$;

$Ts_{i,j,l}$: The setup time of machine $WS_{j,l}$ when job $J_i$ is processed on machine $WS_{j,l}$, $j \in \{2,...,m\}$;

$X$: the quantity of job's properties.

$prop_{x,i}$ the property of job $J_i$, $i \in \{1,...,n\}$ $x \in \{1,...,X\}$;

$Tsp_{j,x}$: the required setup times when there is a change in one property $prop_{x,i}$ of two consecutively processed jobs, $j \in \{1,...,m\}$, $x \in \{1,...,X\}$;

$Nsrv_{x,i,i'}^j$: The relationship between properties of job continuously processed on machine $WS_{j,l}$, $i, i' \in \{1,...,n\}$ and $i \neq i'$, $x \in \{1,...,X\}$;

*2.3 Constraints*

(1) Assumptions

$$At_{i,j,l} = \begin{cases} 1 & \text{Job } J_i \text{ is assigned to be processed on machine } WS_{j,l} \text{ at stage } Oper_j \\ 0 & \text{Job } J_i \text{ isn't assigned to be processed on machine } WS_{j,l} \text{ at stage } Oper_j \end{cases} \quad (1)$$

(2) General constraint of flexible flow shops scheduling

$$\sum_{l=1}^{M_j} At_{i,j,l} = 1 \quad (2)$$

$$C_{i,j} = S_{i,j} + Tb_{i,j}, i \in \{1,2,...,n\}, j \in \{1,2,...,m\} \quad (3)$$

$$C_{i,j-1} \leq S_{i,j}, i \in \{1,2,...,n\}, j \in \{2,2,...,m\} \quad (4)$$

Equation (2) indicates job $J_i$ at stage $Oper_j$ can only be processed on one machine. Eq.(3) constrains the relationship between start time and completion time for job $J_i$ at stage $Oper_j$. Equation (4) represents that job $J_i$ needs to complete the current stage before proceeding to the next stage.

(3) Constraints of limited buffer

$$To_{i,j} = \begin{cases} Te_{i,j+1} & i = \{1,...,m-1\} \\ C_{i,j} & i = m \end{cases} \quad (5)$$

In Eq.(5), when the job is at stage $Oper_j$ the departure time of the job on the machine is equal to the entry time of the job into the buffer. When the job is at stage $Oper_m$, the departure time of the job on the machine equals the completion time.

$$Te_{i,j} \geq C_{i,j-1}, j \in \{2,...,m\} \quad (6)$$

Equation (6) indicates that the entry time of the job into the buffer is greater than or equal to the completion time of the job in the previous stage. If the limited buffer is blocked, the job will be retained on the machine after completing the previous stage.

$$WA_j(t) = \{J_i | OA_{i,j}(t) = 1\} \quad (7)$$

Equation (7) shows all jobs contained in the waiting processing sequence of limited buffer at time $t$.

$$card(WA_j(t)) \leq K_j \quad (8)$$

Equation (8) denotes that at any time, the quantity of jobs in the waiting processing sequence $WA_j$ is less than or equal to the maximum number of machines ($K_j$) in sequence buffer.

(4) Constraints of setup times

$$\sum_{l=1}^{M_j}(Ts_{i,j,l} \cdot At_{i,j,l}) + C_{i,j-1} \leq S_{i,j}, j \in \{2,...,m\} \quad (9)$$

Equation (9) extends the basic constraint (4) of flexible flow shops to obtain the relationship between setup times and start time as well as completion time. Equation (9) indicates that the start time of the current stage is greater than or equal to the completion time of the previous stage plus setup times.

$$Nsrv^l_{x,i,i'} = \begin{cases} 1 & porp_{x,i} \neq porp_{x,i'} \\ 0 & porp_{x,i} = porp_{x,i'} \end{cases} \quad (10)$$

In Eq. (10), $J_i$ and $J_{i'}$ ($i,i' \in \{1,...,n\}$) are jobs continuously processed on machine $WS_{j,l}$. When $Nsrv^l_{x,i,i'}=0$, it means that when property $prop_{x,i}$ of job $Bus_i$ and property $prop_{x,i'}$ of job $Bus_{i'}$ are the same, no setup time is required to process the latter job on the machine. When $Nsrv^l_{x,i,i'}=1$, it represents that when property $prop_{x,i}$ of job $J_i$ and property $prop_{x,i'}$ of job $J_{i'}$ are different, the setup time is required to process the latter job on the machine.

$$Ts_{i',j,l} = \sum_{x=1}^{X} Tsp_{j,x} \cdot Nsrv^l_{x,i,i'}, \quad j \in \{2,...,m\} \quad (11)$$

In Eq. (11), at stage $Oper_j$, $Tsp_{j,x}$ represents the required setup times when there is a change in one property $prop_{x,i}$ of two consecutively processed jobs. $TSP_j = \{Tsp_{j,x}\}$ denotes the collection of setup times when there is a change in the property of two consecutively processed jobs at stage $Oper_j$, and $x \in \{1,...,X\}$. $Ts_{i',j,l}$ is the required setup time when several properties of job $Bus_{i'}$ processed on machine $WS_{j,l}$ at stage $Oper_j$ are different from those of the previous job processed on the same machine.

## 3 Improved whale optimization algorithm

### 3.1 Standard whale optimization algorithm

The bionics principle of the WOA algorithm stems from a special hunting of humpback whales—bubble-net. When the humpback whale finds a prey, it follows a spiral-like path to make a bubble net for foraging. The specific behavior includes three main ways: encircling prey, search for prey, and hunting prey. The foraging behavior of humpback whales can be described as follows:

(1) Encircling prey

The searching process of humpback whales is mainly to encircle the prey by moving about. This behavior is abstracted into the following mathematical model.

$$D = |CX^*(t) - X(t)| \quad (12)$$
$$X(t+1) = X^*(t) - A \cdot D \quad (13)$$

In equations, $t$ represents the current iterations, while A and C denote coefficients. $X^*(t)$ indicates the best position of the humpback whale until the current $t$ generation. $X(t)$ signifies the location of the whale at the moment. The changes of A and C in the whale optimization algorithm are determined by the following way:

$$A = 2a \cdot r_1 - a \quad (14)$$
$$C = 2 \cdot r_2 \quad (15)$$

$r_1$ and $r_2$ are obtained by arbitrarily generating numbers in (0, 1), and the value of $a$ is gained from a regular drop of the iterations from 2 to 0. The equation is as follows:

$$a = 2 - 2t/T_{max} \quad (16)$$

Where $t$ represents the current evolutional iterations and $T_{max}$ is the maximum termination generation.

(2) Search for prey

The mathematical model of search for prey is depicted as follows:

$$D = |CX_{rand} - X(t)| \quad (17)$$
$$X(t+1) = X_{rand} - A \cdot D \quad (18)$$

Where $X_{rand}$ represents any whale's position in a population, which is used to update the current position of the whale. The $|A|$ determines the mode to search for or encircle the prey. When $|A| \geq 1$, the searching mode is adopted; when $|A| \geq 1$, the encircling mode is taken.

(3) Hunting prey

In the process of hunting prey, humpback whale swims to the prey in a spiral motion, so this process can be abstracted into the following mathematical model:

$$D_P = |X^*(t) - X(t)| \quad (19)$$
$$X(t+1) = X^*(t) + D_P \cdot e^{bl} \cdot \cos(2\pi l), l \in (-1,1) \quad (20)$$

Equation (20) represents the distance between the prey and the whale, and $X^*(t)$ indicates the best position of the whale currently. As a constant, b determines the shape of the spiral line.

According to the specific research on the foraging behavior of humpback whales, these three behaviors are synchronous. As such, the mathematical model is shown as follows:

$$X(t+1) = \begin{cases} X^*(t) - AD & p < P_i, |A| < 1 \\ X_{rand} - AD & p < P_i, |A| \geq 1 \\ X^*(t) + D_p e^{bl} \cos(2\pi l), & p \geq P_i \end{cases} \quad (21)$$

In the process, the whales swim around the prey along a spiral-shaped path and within a shrinking circle simultaneously. To model this simultaneous behavior, Mirjalili S assumes that there is a probability of 50% (namely $P_i = 0.5$) to choose between either the shrinking encircling mechanism or the spiral model to update the position of whales during optimization.

### 3.2 Improved whale optimization algorithm

(1) Levy flight

In the original whale optimization algorithm, the ability of the individual to update in the new process is determined by the coefficient value $A$ and the distance between the individual of each generation and the optimal individual. Since value $A$ is a random number in [-2, 2], its range is small and it will become increasingly smaller with the increase of iterations. When all individuals in the population are around the optimal individual and value $A$ is small, it will result in decreasing search range for individual updates and poor update capabilities. Consequently, as iterations increase, the difference between groups becomes smaller and smaller, and the ability to search is weaker and weaker. In this way, the original individual update process is easy to fall into but difficult to jump out of local extremum. Based on the shortcoming that the updating step-size of the WOA algorithm in the evolution becomes smaller due to increasingly smaller individual differences, the Levy step-size is used as the coefficient of coefficient $A$ to simulate the step-size of the whale swimming process. The combination of short-distance and long-distance Levy step-size expands the search range of the WOA algorithm and makes it more powerful in local search and better to jump out of the best local solution.

The way to randomly generate Levy step-size is divided into three parts(MANTEGNA, 1992).

$$s = \frac{u}{|v|^{1/\beta}} \quad (22)$$

Where $u$ and $v$ are given by the normal distribution of equation (23), and $\beta \in [0.3, 1.99]$ is determined by the equation $\Gamma(z)$ which is the gamma function.

$$u \sim N(0, \sigma_u^2) \quad v \sim N(0, \sigma_v^2) \quad (23)$$

$$\sigma_u^2 = \left\{ \frac{\Gamma(1+\beta)\sin(\pi\beta/2)}{\Gamma((1+\beta)/2)\beta \cdot 2^{(\beta-1)/2}} \right\}, \quad \sigma_v = 1 \quad (24)$$

Apply Levy flight to the whale optimization algorithm. Then the encircling prey is:

$$X(t+1) = \left| X^*(t) - Levy(\beta) \cdot D * A \right| \quad (25)$$

The search for prey is:

$$X(t+1) = \left| X_{rand} - Levy(\beta) \cdot D * A \right| \quad (26)$$

(2) Opposition-based learning

As to the WOA algorithm, the individual differences become increasingly smaller with the increase of iterations in the evolutional iterative process of the population. It will make the population fall into local extremum easily in the later stage due to a decrease in the diversity of the population and smaller range of spatial search of the individual. Because the convergence speed of the WOA algorithm is faster, this paper introduces the idea of the opposition-based learning strategy into the update of the individual iteration of the WOA algorithm. By calculating the population congestion degree, namely the density of individuals in the population, we can judge when to start the opposition-based learning to update the individuals.

The population congestion degree variable is set as $\delta$, whose calculation formula is as shown in equation (27).

$$\delta = \sum_{i=1}^{m} \left( \frac{f_i - f_{avg}}{f} \right)^2 \quad (27)$$

Among them, $f_i$ is the fitness function value of the individual; $f_{avg}$ is the average fitness of the population, and $f$ represents the optimal value of the fitness values in the population. The degree of individual aggregation in the whale population is calculated by the above equation. The smaller the $\delta$, the greater the population congestion degree, indicating that much more individuals are clustered near the optimal value. Therefore, the threshold is set to a small value. When , except for 10% individuals with high fitness value, the other individuals based on the principle of opposition-based learning strategy are updated according to the equation (28). After that, the diversity of the population is increased, and the algorithm still has the ability to continuously evolve in the case of decreasing population diversity.

$$X(t+1) = |X_{max} + X_{min} - X^*(t)| \quad (28)$$

(3) Simulated annealing

During the iterations of the whale optimization algorithm as a global optimization algorithm, the individual has been searching in space. After each iteration, the individual will be directly replaced by the new one. In this case, the inferior individuals may be retained. What's more, it is not conducive to maintain population diversity as well as the global search in space. According to the idea of simulated annealing, if the updated individual is better than the original, the updated one will be regarded as the individual of the new population; if the fitness value of the updated individual is worse than that of the original, a selection probability $P$ calculated in accordance with equation (29) is used to accept this inferior individual. In equation (29), $f(i)$ and $f(j)$ are the objective function values of state $i$ and $j$, respectively; $T$ represents the current temperature. This paper applies the idea of simulated annealing that accepts new individuals to the individual update of the whale optimization algorithm, so as to maintain the population diversity in the evolution process.

$$P(i \to j) = \begin{cases} 1, & f(i) > f(j) \\ \exp(\frac{f(i)-f(j)}{T}), & f(i) \leq f(j) \end{cases} \quad (29)$$

(4) Process of improved whale optimization algorithm

Step 1: Initially start the algorithm program, set the number of individuals in the population to NP, and set the maximum termination generation in algorithm optimization to $T_{max}$.

Step 2: According to the coding rule, NP individuals are randomly generated as the initial population. Then the fitness value of all individuals is obtained in accordance with the fitness function of the problem, and the individual with the highest fitness value is taken as the optimal position.

Step 3: At the beginning of the iteration, a probability value P and the coefficient $Levy(\ )$ of the Levy step-size are randomly generated;

Step 4: If P>=0.5, the mode of hunting prey is selected to update the individual, whereas the searching or encircling mode is chosen; if the searching or encircling mode is taken, when A>=1, the mode of search for prey is applied to update the individual; when A<1, the mode of encircling prey is taken.

Step 5: The idea of simulated annealing is employed to accept the updated individual. If the fitness value of the updated individual is better than that of the previous, the updated one will be retained; if the fitness value of the updated individual is less than that of the previous, the new individual will be accepted based on certain probability.

Step 6: Calculate the population congestion degree according to equation (27). If the population density is too small ( ), 10% individuals are retained, and the rest are updated by using the opposition-based learning strategy.

Step 7: Repeat steps 3, 4, 5, and 6 until the maximum generation of the iteration is reached. Then stop the calculation, and output the optimal value.

4  Simulation experiment

Firstly, in order to verify the improvement of the IWOA algorithm on the optimization ability of the standard WOA algorithm, this paper respectively uses examples of four groups of small-scale and four groups of large-scale flexible flow shop scheduling problems (FFSP) to test the IWOA algorithm. Then, the instance data of limited buffer scheduling problems in flexible flow shops with setup times is applied to further verify the effectiveness of the IWOA algorithm for solving such scheduling problems. At the same time, the ICA algorithm and the BA algorithm are used as the comparison algorithm. Both of these two algorithms are new meta-heuristic algorithms. The ICA algorithm was proposed by Atashpaz and Lucas in 2007 and the BA algorithm was proposed by Yang in 2010. At present, these two new algorithms are widely used in scheduling problems and have achieved good results.The algorithm is written by MATLAB 2016a simulation software, running on the

Window10 operating system, with Intel(R) Core(TM) i5-6200U CPU @ 2.30GHz processors, 8GRAM.

*4.1 Optimization performance testing on the algorithm*

The ICA algorithm, BA algorithm, WOA algorithm and IWOA Algorithm are tested using four groups of small-scale and four groups of large-scale FFSP data respectively, so as to verify the optimization effect of the IWOA algorithm. Four groups of small-scale data come from 98 standard examples proposed by Carlier and Neron based on the standard FFSP. The set of examples is divided into five classes(Alaykyran, Engin & Doyen, 2007). And the maximum evolutional generation of four algorithms is set to 500 generations, while the number of the population is set to 30. Each algorithm runs 30 times on each group of data. The average makespan $\overline{C_{max}}$ is used as the foremost evaluation index. The test results are shown in the table below.

**TABLE 1** $\overline{C_{max}}$ obtained by each algorithm under different scale data

| Example | LB | ICA | BA | WOA | IWOA |
|---|---|---|---|---|---|
| j10c10c3 | 98 | 118.93 | 117.40 | 117.40 | 116.87 |
| j10c10c4 | 103 | 121.03 | 119.77 | 119.77 | 118.97 |
| j15c5d3 | 77 | 86.60 | 85.23 | 85.23 | 83.73 |
| j15c5d4 | 61 | 88.03 | 86.97 | 86.97 | 84.27 |
| j80c8d1 | — | 1856.07 | 1828.63 | 1828.63 | 1789.60 |
| j80c8d2 | — | 1864.57 | 1844.97 | 1844.97 | 1816.77 |
| j120c4d1 | — | 2173.57 | 2146.23 | 2146.23 | 2114.40 |
| j120c4d2 | — | 2290.10 | 2259.40 | 2259.40 | 2219.17 |

In the table, *LB* represents the lower bound of makespan for the examples proposed by Carlier and Neron, whose optimal value has been given by Santos D L (1995) and Nerson E (2001). It can be seen from the table that the IWOA algorithm is better than the other three algorithms in the test of four groups of small-scale standard examples. Especially for the two examples (j15c5d3 and j15c5d4) of the d-class problem, the optimization effect of the IWOA algorithm is significantly better than those of the other three algorithms. For the example j15c5d3, the average relative error (*ARE*) between the solution obtained by the IWOA algorithm and the lower bound of makespan is 8.74%, which is much smaller than 12.47% of the ICA algorithm, 10.69% of the BA algorithm and 11.26% of the WOA algorithm. For the example j15c5d4, the *ARE* of these four algorithms is 44.31%, 42.57%, 41.52%, 38.15% respectively, where *ARE* of IWOA algorithm is also the smallest. As to the four groups of large-scale data, the optimization result obtained by the IWOA algorithm has been greatly improved compared with those of the other. Among them, for the j80c8d class problem, the average values of the IWOA, ICA, BA and WOA algorithms are 1803.19, 1860.32, 1836.8, and 1827.84 respectively. For the j120c4d class problem, the average values of the IWOA, ICA, BA and WOA algorithms are 2166.79, 2231.84, 2202.82, and 2191.68 respectively.

According to the above analysis, the IWOA algorithm has better effects than the ICA, BA and WOA algorithms, whether it is to solve small-scale or large-scale data scheduling problems. This shows that compared with the other three algorithms, the IWOA algorithm has stronger abilities to continuously evolve and jump out of local extremum, which overcomes the premature convergence of WOA algorithm to some extent.

*4.2 Instance test on limited buffer in flexible flow shops with setup times*

*4.2.1 Establishing simulation data*

The simulation data of the production operations in the welding and painting workshops of the bus manufacturer were established as follows (Lee, Lai & Roan, 2017).
(1) Parameters in the workshop model
The welding shop of the bus manufacturer has one production stage, and the production of painting workshop has three stages. Therefore, setting the simulation data is set to include four stages, namely $\{Oper_1, Oper_2, Oper_3, Oper_4\}$, whose parallel machine $\{M_j\}$ is $\{3,2,2,2\}$. The buffers among four stages all have limited buffers, and the capacity of each limited buffer is $\{2,2,1\}$.

In the production of the painting workshop, it is necessary to clean the machine and to adjust production equipment if the model and color of the buses that are successively processed on the machine are different. Therefore, the simulation process uses the changes in the model and color as the basis for calculating setup times. Table 2 shows that the setup time parameters are set when the model and color of the buses that are processed successively on the machine have changed. When the bus is assigned to the machine of the next stage from the buffer, the setup time of the machine is calculated using Eq. (11).

**TABLE 2** Parameters of setup times

| | parameter | Description | volue |
|---|---|---|---|
| parameter of setup times | $Tsp_{2,1}$ | Setup time parameters when the model of buses processed successively on a parallel machine of stage $Oper_2$ changes | 3 |
| | $Tsp_{2,2}$ | Setup time parameters when the color of buses processed successively on a parallel machine of stage $Oper_2$ changes | 3 |
| | $Tsp_{3,1}$ | Setup time parameters when the model of buses processed successively on a parallel machine of stage $Oper_3$ changes | 2 |
| | $Tsp_{3,2}$ | Setup time parameters when the color of buses processed successively on a parallel machine of stage $Oper_3$ changes | 2 |
| | $Tsp_{4,1}$ | Setup time parameters when the model of buses processed successively on a parallel machine of stage $Oper_4$ changes | 2 |
| | $Tsp_{4,2}$ | Setup time parameters when the color of buses processed successively on a parallel machine of stage $Oper_4$ changes | 2 |

(2) Parameters of processing object

The sum of bus properties is 2, namely $X=2$. $Prop_1$ represents the model property of the bus, while $Prop_2$ denotes the color property of the bus. The value of model property ($PropValue_1$) is $\{Type_1, Type_2, Type_3\}$, and the value of color property ($PropValue_2$) is $\{Color_1, Color_2, Color_3\}$.

**TABLE 3** Information of bus properties

| property | Model $Prop_1$ | Color $Prop_2$ |
|---|---|---|
| $J_1$ | $Type_1$ | $Color_3$ |
| $J_2$ | $Type_1$ | $Color_3$ |
| $J_3$ | $Type_2$ | $Color_1$ |
| $J_4$ | $Type_1$ | $Color_3$ |
| $J_5$ | $Type_1$ | $Color_3$ |
| $J_6$ | $Type_3$ | $Color_3$ |
| $J_7$ | $Type_1$ | $Color_3$ |
| $J_8$ | $Type_1$ | $Color_3$ |
| $J_9$ | $Type_2$ | $Color_1$ |
| $J_{10}$ | $Type_1$ | $Color_3$ |
| $J_{11}$ | $Type_3$ | $Color_3$ |
| $J_{12}$ | $Type_2$ | $Color_1$ |
| $J_{13}$ | $Type_1$ | $Color_1$ |
| $J_{14}$ | $Type_3$ | $Color_3$ |
| $J_{15}$ | $Type_2$ | $Color_1$ |

(3) Processing time for the bus

**TABLE 4** Standard processing time for bus production

| $Oper$ | $Oper_1$ | $Oper_2$ | $Oper_3$ | $Oper_4$ |
|---|---|---|---|---|
| $J_1$ | 28 | 22 | 14 | 12 |
| $J_2$ | 31 | 18 | 16 | 20 |
| $J_3$ | 35 | 18 | 24 | 26 |
| $J_4$ | 39 | 15 | 12 | 22 |
| $J_5$ | 30 | 19 | 22 | 14 |
| $J_6$ | 35 | 21 | 20 | 33 |
| $J_7$ | 42 | 15 | 18 | 18 |
| $J_8$ | 31 | 23 | 24 | 22 |
| $J_9$ | 30 | 16 | 12 | 13 |
| $J_{10}$ | 30 | 16 | 25 | 22 |
| $J_{11}$ | 42 | 12 | 24 | 18 |
| $J_{12}$ | 31 | 22 | 16 | 33 |
| $J_{13}$ | 20 | 25 | 15 | 14 |
| $J_{14}$ | 23 | 12 | 16 | 30 |
| $J_{15}$ | 34 | 18 | 20 | 18 |

### 4.2.2 Simulation scheme and algorithm parameter table

In order to verify the effectiveness of the IWOA algorithm proposed in this paper for solving the limited buffer scheduling problems in flexible flow shops with setup times, four groups of simulation schemes are set up for comparative analysis. These simulation schemes take the ICA, BA, WOA and IWOA algorithms as the global optimization algorithm respectively, and apply them to solve the above-mentioned scheduling problem of bus manufacturer. The information of the four simulation schemes and the parameter settings of the four global optimization algorithms are shown in Table 5.

**TABLE 5** Simulation scheme and algorithm parameter

| Simulation | Global Optimization algorithm | algorithm parameters |
|---|---|---|
| Scheme 1 | ICA | Maximum training iterations $Gen$ =300; Number of population $NP$ =30; Colonial influence factor =0.15; Similarity threshold $Rt$=0.3; |
| Scheme 2 | BA | Maximum training iterations $Gen$ =300; Number of population $NP$ =30; Random factor =0.9; |
| Scheme 3 | WOA | Maximum training iterations $Gen$ =300; Number of population $NP$ =30; Selection probability $P_t$ =0.5; |
| Scheme 4 | IWOA | Maximum training iterations $Gen$ =300; Number of population $NP$ =30; Selection probability $P_t$ =0.5; |

### 4.2.3 Simulation results and analysis

(1) Evaluation index of scheduling results

In the optimization process, makespan $C_{max}$ is used as the fitness function value of the global optimization algorithm. Meanwhile, a number of evaluation indexes related to the actual production line are established, including the waiting processing time for total job $TWIP$, the total machine setup times $TS$ and the total job blocking time $TPB$ (Jiang & Pan, 2018).

Makespan

$$C_{max} = \max\{C_{i,j}\}, i \in \{1,2,..n\}, j \in \{1,2,..m\} \quad (30)$$

In Eq. (30), $C_{max}$ indicates the maximum completion time of all jobs processed at the last stage, which is also the time for all jobs to complete the process.

Waiting processing time for total

$$TWIP = \sum_{j=2}^{m}\sum_{i=1}^{n}(S_{ij} - C_{i,j-1}) \quad (31)$$

In Eq. (31), the waiting time of the job is from the completion moment $C_{i,j-1}$ of job $Bus_i$ at stage $Oper_{j-1}$ to the start moment $S_{ij}$ at the next stage $Oper_j$. $TWIP$ represents the sum of waiting time of all jobs processed in the entire production. In the production workshop of limited buffer, the waiting time of each job between stages is equal to the sum of time for the job staying on the buffer ($Tl_{i,j} - Te_{i,j}$), blocking time of the job on the machine ($Te_{i,j} - C_{i,j-1}$), and setup times $\sum_{l=1}^{M_j}(Ts_{i,j,l} \cdot At_{i,j,l})$.

Total machine setup times

$$TS = \sum_{i=1}^{n}\sum_{j=1}^{m}\sum_{l=1}^{M_j}(Ts_{i,j,l} \cdot At_{i,j,l}) \quad (32)$$

In Eq. (32), $TS$ is the sum of all stages' setup times in flexible flow shops.

Total job blocking time

$$TPB = \sum_{i=1}^{n}\sum_{j=2}^{m}(Te_{i,j} - C_{i,j-1}) \quad (33)$$

In Eq. (33), $TPB$ is the sum of blocking time of all jobs stuck on the machine due to the buffer being full after all jobs finish the processing in flexible flow shops (Han et al.,2017).

Four sets of simulation schemes run 30 times respectively. The optimal value, the worst value, and the average value of the 30 simulation results are recorded in Table 6. It can be seen from Table 6 that the average of four evaluation indexes of the scheme 4 is the smallest among the four schemes, which indicates that the optimization effect of the IWOA algorithm on four evaluation indexes is better than those of the other three algorithms. In particular, for evaluation indexes $TWIP$ and $TPB$, the optimization effect of the IWOA algorithm is significantly better than those of the other. Compared with the other three schemes, the $TWIP$ of scheme 4 with the IWOA algorithm is reduced by 16.44%、12.85% and 7.37%, and the $TPB$ is reduced by 64.33%、55.30% and 48.38%.

**TABLE 6** Evaluation index comparison of scheduling results of four schemes

| Evaluation index | | Scheme 1 | Scheme 2 | Scheme 3 | Scheme 4 |
|---|---|---|---|---|---|
| $C_{max}$ | Optimum | 230 | 229 | 230 | 226 |
| | Worst | 244 | 238 | 239 | 234 |
| | Average | 236.33 | 233.58 | 234.42 | 230.79 |
| $TWIP$ | Optimum | 76 | 83 | 77 | 63 |
| | Worst | 153 | 141 | 134 | 120 |
| | Average | 118.79 | 113.89 | 107.16 | 99.26 |
| $TS$ | Optimum | 34 | 32 | 37 | 32 |
| | Worst | 80 | 82 | 81 | 75 |
| | Average | 60.79 | 58.05 | 58.78 | 54.63 |
| $TPB$ | Optimum | 2 | 1 | 0 | 0 |
| | Worst | 19 | 19 | 17 | 12 |
| | Average | 10.29 | 8.21 | 7.11 | 3.67 |

(2) Gantt chart analysis of scheduling result

Figure 2 is the Gantt chart of scheduling result of scheme 4. The green part indicates the residence time of the bus in the buffer. The red part indicates the setup time when the successively-processed buses are with different properties. The blue part denotes the blocking time of the bus on the machine after the completion of the process. The processing route of $J_{12}$ is $\{WS_{1,2}, WS_{2,2}, WS_{3,1}, Bs_{4,1}, WS_{4,1}\}$. In Fig. 2, we can see that at time $t=104$, $J_{12}$ completes the processing on machine $WS_{3,1}$ of stage $Oper_3$ (the competition time of $J_{12}$ is 104, namely $C_{12,3}=44$). At this time, $WA_4(t)=\{J_{11}\}$, $card(WA_4(t))=1$, and the capacity of buffer $Bu_4$ is 1 ($K_4=1$); $card(WA_4(t)) \le K_4$ satisfies the constraint (8) of the limited buffer. After being processed at stage $Oper_3$, $J_{12}$ is blocked on machine $WS_{3,1}$ of stage $Oper_3$. Until at time $t=109$ when $J_6$ completes the processing of stage $Oper_4$ and then $J_{11}$ is assigned to the machine $WS_{4,2}$ of stage $Oper_4$, do $J_{12}$ enter buffer $Bu_4$ of stage $Oper_4$ and wait for processing. That is, the time for $J_{12}$ to enter buffer $Bu_4$ is 109 (namely, $Te_{12,4}$ = 109). $Te_{12,4}$ $C_{12,3}$ meets the limited buffer constraint (6).

**Figure 2**  Gantt diagram of IWOA algorithm

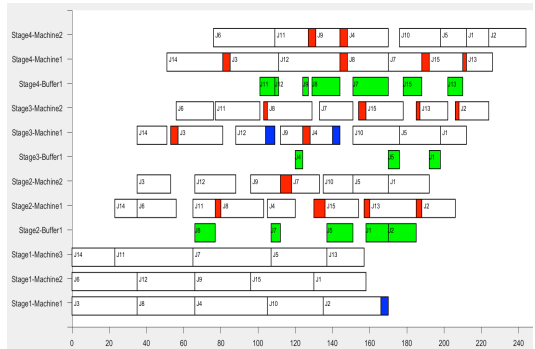

(3) Analysis of scheduling evolution

As can be seen from Figure 3, the ICA algorithm converges very fast in the initial stage of evolution but stops evolving in the 49th generation. What's more, its optimization effect is also the worst among the four algorithms, which reflects that the ICA algorithm evolves quickly but easily falls into local extremum. Although the optimization effect of the BA and WOA algorithms is superior to that of the ICA algorithm, they also stop evolving in the 112th and 96th generations respectively, and fall into local extremum. Since the IWOA algorithm converges slowly in the initial stage, it does not stop evolving until the 152th generation, whose optimization effect is the best among the four algorithms. When the IWOA algorithm shows evolutionary stagnation in the 42nd generation, it jumps out of local extremum and continues to evolve forward under the joint action of Levy flight, opposition-based learning strategy and simulated annealing.

**Figure 3**  Evolution curve of 4 algorithms

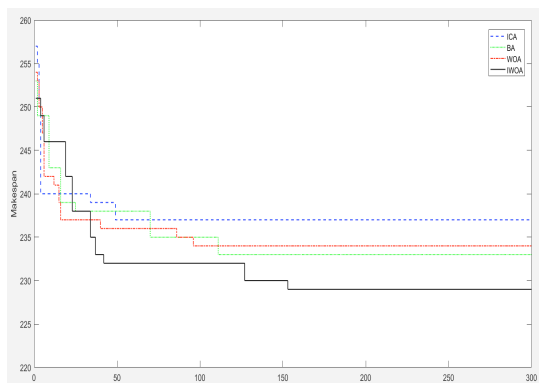

## 5 Conclusion

This paper investigates the limited buffer scheduling problems in flexible flow shops with setup times and applies the whale optimization algorithm as the global searching algorithm. To avoid falling into local extremum, this paper improves WOA by introducing Levy flight, opposition-based learning strategy and simulated annealing method to expand the search range of the algorithm and improve the evolution capability of the algorithm. Based on the verification data of welding and painting workshops of bus manufacturer, simulation tests are made on the ICA, BA, WOA and IWOA algorithms, and a number of evaluation indexes are set to evaluate the optimization results of these algorithms. The simulation results show that compared with the other state of the art algorithms, the IWOA algorithm can better solve this type of limited buffer scheduling problem in flexible flow shops with setup times. For future work, WOA will be improved for better performance in the cloud-computing environment, by exploring the computing power and storage capacity of the cloud computing platform.

**Acknowledgement**

This work was supported by Liaoning Provincial Science Foundation (No. 201602608, 201602615), Natural Science Foundation of China (No. 61873174, 61503259) , Project of Liaoning Province Education Department (LJZ2017015) and Shenyang Municipal Science and Technology Project(No. Z18-5-015).